\documentclass{ecai}
\usepackage{graphicx}
\usepackage{latexsym}
\usepackage{multirow}
\usepackage{inconsolata}
\usepackage{subfigure}
\usepackage[caption=false]{subfig}
\usepackage{pgfplots} 
\usepackage{adjustbox}
\usepackage{arydshln}
\usepackage{stfloats}
\begin{document}

\begin{frontmatter}

\title{
Multilingual Lexical Simplification via Paraphrase Generation}
\author[A]{\fnms{Kang}~\snm{Liu}}
\author[A]{\fnms{Jipeng}~\snm{Qiang}\thanks{Corresponding Author. Email: jpqiang@yzu.edu.cn}}
\author[A]{\fnms{Yun}~\snm{Li}}
\author[A]{\fnms{Yunhao}~\snm{Yuan}}
\author[A]{\fnms{Yi}~\snm{Zhu}}
\author[B]{\fnms{Kaixun}~\snm{Hua}} 

\address[A]{School of Information Engineering, Yangzhou University, Yangzhou, China}
\address[B]{Department of Industrial and Management Systems Engineering, University of South Florida, Tampa, United States}

\begin{abstract}
Lexical simplification (LS) methods based on pretrained language models have made remarkable progress, generating potential substitutes for a complex word through analysis of its contextual surroundings. However, these methods require separate pretrained models for different languages and disregard the preservation of sentence meaning. In this paper, we propose a novel multilingual LS method via paraphrase generation, as paraphrases provide diversity in word selection while preserving the sentence's meaning. We regard paraphrasing as a zero-shot translation task within multilingual neural machine translation that supports hundreds of languages. After feeding the input sentence into the encoder of paraphrase modeling, we generate the substitutes based on a novel decoding strategy that concentrates solely on the lexical variations of the complex word. Experimental results demonstrate that our approach surpasses BERT-based methods and zero-shot GPT3-based method significantly on English, Spanish, and Portuguese.
\end{abstract}

\end{frontmatter}

\section{Introduction}
Lexical Simplification (LS) \cite{paetzold2017survey} aims to replace complex words in sentence with simple alternatives while keeping the original sentence meaning, which is a key task to facilitate reading comprehension for different target readerships such as non-native speaker\cite{paetzold2016unsupervised}, people with cognitive disabilities\cite{saggion2017automatic}. Earlier LS methods are primarily rule-based\cite{carroll1998practical,biran2011putting} or relied on word embedding models\cite{glavavs2015simplifying,paetzold2016unsupervised}. However, such approaches only account for individual complex word, resulting in many candidate substitutes that do not fit for the context. In recent years, LS methods based on pretrained language models including BERT and its variations \cite{qiang2021lsbert,li-etal-2022-mantis,north-etal-2022-gmu,whistely-etal-2022-presiuniv} generate substitutes by predicting the probability distribution of the vocabulary from the representation of the complex word based on its context, and have emerged as the predominant technique compared with previous LS methods. But, there remains two limitations for them:

(1) Poor multilingual scalability. Considering the available pretrained language models, such efforts have coalesced around a small subset of languages, leaving behind the vast majority of mostly low-resource languages. Additionally, for LS tasks in different languages, it is necessary to use pretrained language models in different languages, which greatly limits the effectiveness and applicability of this type of method in a multilingual environment. 

(2) Disregarding the preservation of sentence meaning.   The generated substitutions are both semantically coherent with the complex word and contextually appropriate. But, there is no guarantee that the generated substitutions could still preserve the original sentence's meaning \cite{lin-etal-2022-improving,wada-etal-2022-unsupervised}. For example, given one sentence "Tom is a bad guy", the substitutes for word "bad" using BERT are "good, rough, big, dangerous".

To address those limitations mentioned above, we study how to generate simpler substitutes for complex word via paraphrase modeling. (1) Inspired by one work \cite{thompson2020paraphrase}, we use a paraphraser via multilingual Neural Machine Translation (NMT) system (NLLB) based on encoder-decoder framework\cite{costa2022no}, enabling high-quality zero-shot translations in 200 languages. By configuring the output language to correspond with the input language, multilingual NMT can generate paraphrases directly, overcoming the first limitation. (2) Paraphrases generated from the paraphraser provide diversity in word selection while preserving the sentence's meaning\cite{hu2019parabank,hao2022parazh}. The meaning-preserving properties of paraphrase models can aid in addressing the second limitation. But, no studies have been conducted to utilize paraphraser to generate the substitutes. Because the output paraphrases using the existing decoding strategy concern the lexical variations of the whole sentence rather than the complex word, it becomes challenging to extract substitutes from them. Therefore, the big challenge we face is how to generate the paraphrases that only concerns the lexical variations of the complex word, rather than the whole sentence.

In this paper, we propose a multilingual LS method via multilingual NMT, which adopts one novel decoding strategy that focuses on lexical variations of the complex word. We first force the decoder begin with the complex word's prefix, to subsequently generate the probability distribution of the complex word's position. Then, we adopt a re-scoring approach that incorporates an estimate of the complex word's suffix to make a more knowledgeable choice. By following this methodology, the generated paraphrases only concern the lexical variations of the complex word.

Our primary contributions in this paper are as follows:

(1) We are the first to introduce the idea of utilizing a multilingual NMT to tackle the challenge of LS. Our method guarantees that the generated substitutions can surely better preserve the original sentence meaning. Moreover, our approach relies on a single model and accommodates various languages.

(2) We propose a simple but effective decoding strategy for substitution generation. Our decoding strategy can effectively identify the lexical variations of complex word and select candidate words that are most suitable for the given context. 

(3) Experimental results on the TSAR-2022 multilingual LS benchmark (English, Spanish and Portuguese), our zero-shot method outperforms previous BERT-based methods and zero-shot GPT3-based method by a significant margin, and shows marginal improvements over the ensemble GPT3-based method. We release our code and the results at github \footnote{https://github.com/KpKqwq/LSPG}.

\section{Related work}
\textbf{Lexical Simplification:} LS generally consists of three or four steps: complex word identification, substitution generation, substitution selection (optional), and substitution ranking\cite{paetzold2017survey}. Complex word identification is usually treated as an independent task, which is not addressed in this paper.
Earlier LS methods were rule-based\cite{carroll1998practical,biran2011putting} or relied on word embedding models\cite{glavavs2015simplifying,paetzold2016unsupervised}. These methods typically seek to find synonyms or words similar with the complex word. However, as these methods only take into account individual complex words, they often generate many potential substitutions that are not fit for the context.

Some work \cite{pavlick2016simple, Kriz2018Simplification} utilize large-scale paraphrase rule database, e.g., PPDB \cite{Ganitkevitch2013} or its variations \cite{pavlick2015ppdb,pavlick2016simple,qiang2021unsupervised}, to find substitute candidates for complex words, where the paraphrase rule database consists of large-scale lexical paraphrase rules (e.g., "berries $\rightarrow$ strawberries") that are extracted from large-scale paraphrase sentence pairs, such as ParaNMT \cite{wieting2017paranmt} or ParaBank \cite{hu-etal-2019-large}. These works do not take into account the context as rule-based LS methods do. 

LS methods based on the pretrained model BERT have recently attracted much attention \cite{qiang2021lsbert,li-etal-2022-mantis,north-etal-2022-gmu,whistely-etal-2022-presiuniv}. Such methods involve masking the complex word and predicting potential substitutions based on the context. BERT-based methods have demonstrated significant performance enhancements compared to previous methods, and have now become the dominant approach for LS.

While previous research primarily focused on English, recent advancements in multilingual and cross-lingual NLP models have facilitated studies in other languages\cite{finnimore2019strong,vstajner2022lexical,qiang2021chinese}. This trend is reflected in the Text Simplification, Accessibility, and Readability (TSAR-2022) shared task\cite{saggion2023findings}, which provides participants with multilingual LS datasets in three language tracks: English, Spanish, and Portuguese. The shared task garnered significant interest, with a total of 60 systems submitted across different languages. During the study, participants introduced a range of language-specific BERT-based methods\cite{li-etal-2022-mantis,north-etal-2022-gmu,whistely-etal-2022-presiuniv}. Additionally, Aumiller and Gertz\cite{aumiller-gertz-2022-unihd} submitted two systems based on GPT3, demonstrating the potential of large language models in multilingual LS. However, this method solely relies on paid inference for research purposes. In contrast to the aforementioned methods, we employ a multilingual NMT to tackle the challenge of multilingual LS.

\textbf{Multilingual NMT:} Multilingual NMT has emerged as a rapidly growing field in recent years\cite{dabre2020survey}. Google's multilingual NMT system\cite{johnson2017google} has demonstrated the ability to translate between languages without direct parallel data, which is called zero-shot translation. Multilingual NMT system NLLB \cite{costa2022no} can support translation in over two hundred languages. To enhance the performance of multilingual NMT, researchers have explored various approaches, such as language clustering\cite{tan-etal-2019-multilingual} and multilingual pretraining\cite{fan2021beyond}. 

The potential of multilingual NMT has been explored in several studies, including text similarity measure\cite{vamvas-sennrich-2022-nmtscore} and paraphrase generation\cite{thompson2020paraphrase}. These investigations provide valuable insights into the efficacy of multilingual NMT for diverse applications. However, to our knowledge, our study is the first to examine the utilization of multilingual models for multilingual LS.

\textbf{Decoding method:} In recent years, autoregressive modeling has emerged as a popular approach for text generation tasks such as machine translation\cite{dabre2020survey} and paraphrase generation\cite{zhou2021paraphrase,qiang2023natural}. During inference, beam search is commonly used as the decoding strategy, which involves keeping a fixed number of the most probable partial sequences at each decoding step. Although the basic beam search algorithm is effective, several modifications have been proposed to further enhance its performance for specific objectives. For instance, diverse beam search\cite{vijayakumar2018diverse} aims to improve both diversity and quality of generated sequences. Additionally, for constrained text generation, Lu\cite{lu-etal-2022-neurologic} employs lookahead heuristics to steer the generations towards sequences that satisfy the given constraints. In contrast to these decoding strategies, our approach focuses on identifying substitutions for single complex word.

\section{Method}
Given one sentence $\mathbf{x}=\{x_0,...,x_c,...,x_n\}$ and the complex word $x_c$, we will introduce how to utilize multilingual machine translation system for generating suitable substitutions for $x_c$ in many languages.

 During decoding, we propose an effective decoding strategy to generate substitutes of the complex word. Then, we rank the generated substitutions via three features to select the most appropriate simpler one.

\textbf{Paraphraser:}  By aligning the output language with the input language (e.g., “translation” from English to English), multilingual NMT could be treated as a paraphraser that supports the paraphrasing of multiple languages.  In our method, we use the multilingual NMT system NLLB \cite{costa2022no} that enables high-quality zero-shot translations in 200 languages. NLLB is a standard left to right autoregressive model:$p_{\theta}(\mathbf{y}|\mathbf{x})=\prod_{c=0}^{|\mathbf{y}|}p_{\theta}(y_c|\mathbf{y}_{<c},\mathbf{x})$, trained on different language directions.

\begin{figure}[h!]
    \centering
     \includegraphics[width=\linewidth]{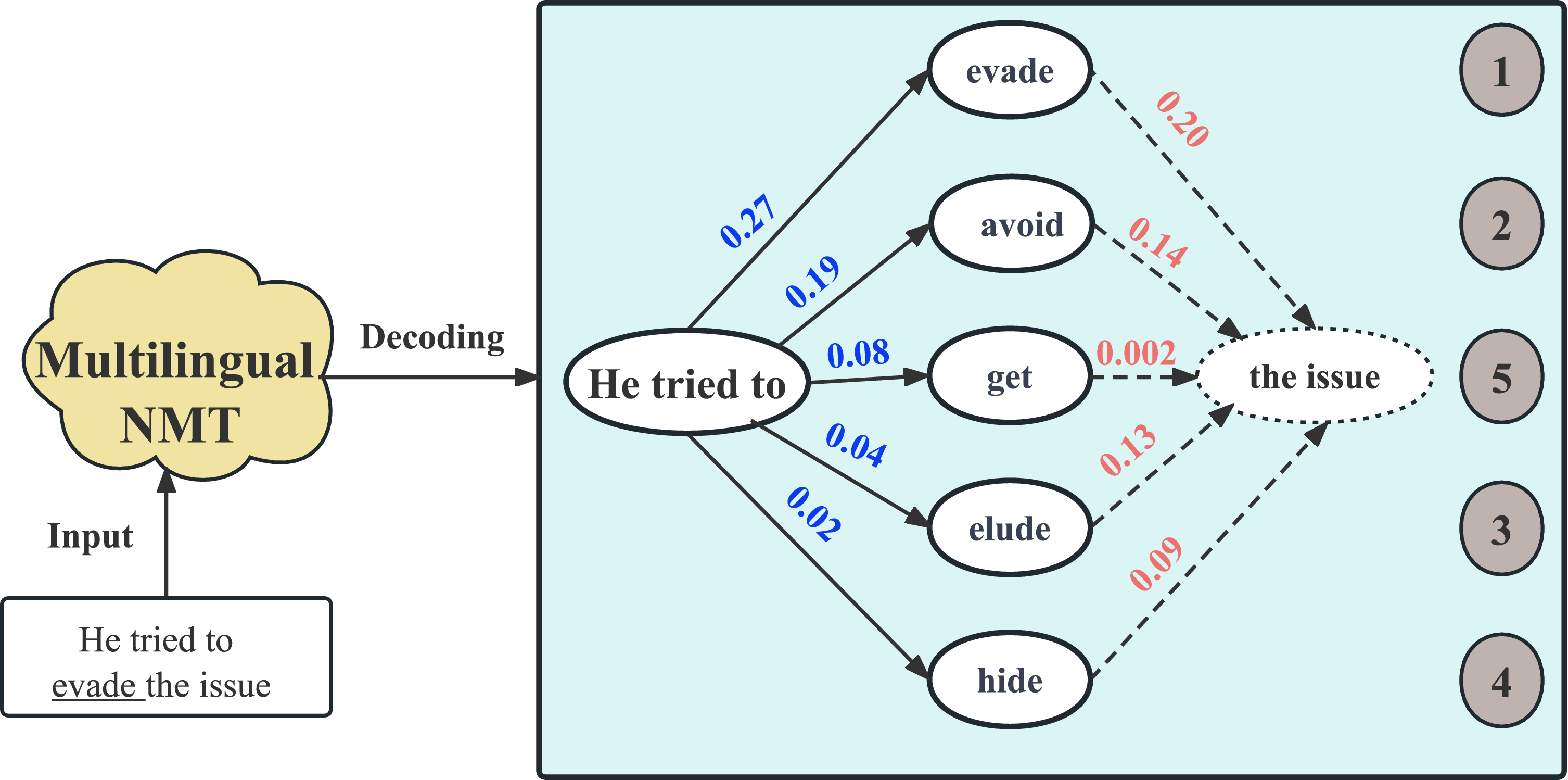} 
\caption{
Substitute generation of our method. Let us consider the example of a sentence "He attempted to evade the issue" containing the complex word "evade". By inputting the sentence into the Multilingual NMT encoder, we derive the top five possible candidates, along with their respective probability scores, by directing the decoder to initiate with the prefix phrase "He attempted to" of the complex word. Subsequently, we re-score these candidates by computing the likelihood of generating suffix words for these candidates. Finally, the top three substitutes, "avoid, elude, and hide", are obtained by eliminating the original word.
% The progress of generating first five paraphrases of `He tried to evade the issue'. The complex word is `evade'. The initial probability values for each candidate are shown in blue, while the red values indicates the estimate scores after factoring in future suffixes.
% Generated paraphrases of "The man's rhetoric was full of lies", rheoric is the complex word. Figure (a) shows the first five generated paraphrase using beam search and Figure (b) shows the
% first 5 paraphrases of our decoding m ethod by focusing the lexical variation of the complex word while taking context into account.
}
\label{fig:framework}
\end{figure}

\textbf{Substitution generation:} The process of substitution generation is employed to produce the candidates of $x_c$. Following the input of $\mathbf{x}$ into the Paraphraser, we can generate a multitude of paraphrases through beam searching. However, it proves to be arduous to extract substitutes from them, as the paraphrases pertain to the lexical variations of the entire sentence instead of the intricate word.

Here, we propose a novel decoding strategy that is exclusively designed to leverage lexical variations of the complex word, as shows in Figure \ref{fig:framework}. After feeding the sentence $\mathbf{x}$ into the encoder of Multilingual NMT, we force the decoder to begin with prefix $\mathbf{x}_{<c}$ of complex word, and decode succeeding token distribution $p_{\theta}(y_c|\mathbf{y}_{<c}= \mathbf{x}_{<c}, \mathbf{x})$. We select the top $K$ tokens $Y_c$ with the highest probability in the distribution as the results of decoding.

\begin{equation}
Y_c = \mathop{\arg topK}_{y_c} \{\log p_{\theta}(y_c|\mathbf{y}_{<c},x)\}
 \end{equation}
where $\mathbf{y}_{<c}=\mathbf{x}_{<c}$

Based on Equation (1), if we directly select the top $K$ tokens $Y_{c}$ with highest probability in token distribution at complex word's position , we could not ensure that the selected candidates are also suitable for the original suffix of the complex word. As Figure \ref{fig:framework} shows, the candidate `get' owning a higher probability is clearly not the variation of `evade', even though it is semantically coherent with the complex word and contextually appropriate.

Inspired by Lu\cite{lu-etal-2022-neurologic}, we incorporate an extended estimate for the original suffix into our scoring function, replacing Equation (1) with:
\begin{equation}
Y_c = \mathop{\arg topK}_{y_c} \{\log p_{\theta}(y_c|\mathbf{y}_{<c},x)+\log p_{\theta}(\mathbf{y}_{>c}|\mathbf{y}_{<c}, y_c,x)\}
 \end{equation}
where $\mathbf{y}_{<c}=\mathbf{x}_{<c}$, $\mathbf{y}_{>c}=\mathbf{x}_{>c}$.

The primary enhancement involves the incorporation of a lookahead heuristic that modifies the score of a candidate $(y_{<c},y_c)$ based on the probability of satisfying additional suffix constraints $y_{>c}$. In practice, it is not necessary to analyze the entire suffix, examining just two or three words is sufficient. As illustrated in Figure 1, the top substitutes generated by our decoding strategy effectively align with the context and successfully retain the intended meaning of the sentence.

\textbf{Substitution ranking:} The step of substitute ranking aims to rank the generated substitutes by their simplicity, which is a necessary step in LS task. Word frequency feature calculated by large corpus is often used to measure the complexity of the substitutes \cite{paetzold2017survey,qiang2020AAAI,saggion2023findings}. Considering that the higher the word frequency, the more simple the word is, this phenomenon could be beneficial to lexical simplification.

We give one simple ranking method that uses three features with different weights to rank the generated substitutes: (1) prediction feature using Equation (2) (the probability of the candidate extracted during the substitute generation), (2) Word Frequency, and (3) semantic similarity (cosine similarity between the word embedding vector of the complex word and the candidate). To support a wide range of languages, we utilize fastText\footnote{https://fasttext.cc/docs/en/crawl-vectors.html} to obtain word embedding vectors supporting 157 languages, and the wordfreq package\footnote{https://pypi.org/project/wordfreq/} to calculate frequency scores for 44 languages. The final score for each substitute is calculated as a weighted sum of the three features.

However, it is imperative to emphasize that with our methodology, the generated substitutes can be utilized directly without the need for substitute ranking. This is because NMT models usually tend to generate more high-frequency tokens and less low-frequency tokens \cite{gu-etal-2020-token,jiang2019improving}. In our experiments, it verifies that our method without substitution ranking has also yielded excellent results.

\section{Experiments}

\subsection{Experiment Setup}

\textbf{Evaluation Datasets:} For reliable comparison of methods' performances across the different languages, we use the newest multilingual LS evaluation datasets from TSAR-2022 shared task\cite{saggion2023findings}, which are composed of three language tracks: English, Spanish and Portuguese. Table 1 presents the dataset statistics, and each dataset is further divided into separate test and validation sets.
\begin{table}
    \centering
    \begin{tabular}{ccccc}\hline
        \multirow{2}{*}{dataset}&\multirow{2}{*}{instances}&\multicolumn{3}{c}{Substitution per target}\\
        &&Min&Max&Avg\\\hline
        EN&386&2&22&10.55\\
        ES&381&2&19&10.28\\
        PT&386&1&16&8.10\\\hline
    \end{tabular}
    \caption{Statistics on the TSAR-2022 share task multilingual LS dataset.}
    \label{tab:statistic}
\end{table}

\begin{table*}[]
    \centering
\resizebox{\linewidth}{!}{
    \begin{tabular}{l|l|c|ccc|ccc|ccc}\hline
          &&&\multicolumn{3}{c|}{Acc@k@Top1} &\multicolumn{3}{c|}{MAP@k}&\multicolumn{3}{c}{Potential@k} \\\hline
          Language&Method&ACC@1&k=1&k=2&k=3&k=3&k=5&k=10&k=3&k=5&k=10\\\hline
          \multirow{6}{*}{English}&GPT3(Ensemble)&0.8096&0.4289&0.6112&0.6863&0.5834&0.4491&0.2812&\textbf{0.9624}&0.9812&0.9946\\
          &GPT3(Single)&0.7721&0.4262&0.5335&0.5710&0.5096&0.3653&0.2092&0.8900&0.9302&0.9436\\\cdashline{2-12} 
            &LSBERT&0.5978&0.3029&0.4450&0.5308&0.4079&0.2957&0.1755&0.8230&0.8766&0.9463\\
            &MANTIS&0.6568&0.3029&0.4450&0.5388&0.4730&0.3599&0.2193&0.8766&0.9463&0.9785\\\cdashline{2-12} 
            &LSPG(w/o ranking) &0.7640&0.4021&0.5656&0.6514&0.5655&0.4351&0.2829&0.9436&0.9839&0.9973\\
            &LSPG&\textbf{0.8176}&\textbf{0.4557}&\textbf{0.6166}&\textbf{0.6890}&\textbf{0.5881}&\textbf{0.4632}&\textbf{0.2994}&\textbf{0.9624}&\textbf{0.9839}&\textbf{0.9973}\\\hline
            
          \multirow{6}{*}{Spanish}&GPT3(Ensemble)&0.6521&0.3505&0.5108&0.5788&0.4281&0.3239&0.1967&0.8206&0.8885&\textbf{0.9402}\\
          &GPT3(Single)&0.5706&0.3070&0.3967&0.4510&0.3526&0.2449&0.1376&0.6902&0.7146&0.7445\\\cdashline{2-12} 
            &LSBERT&0.2880&0.0951&0.1440&0.1820&0.1868&0.1346&0.0795&0.4945&0.6114&0.7472\\
            &PresiUniv&0.3695&0.2038&0.2771&0.3288&0.2145&0.1499&0.0832&0.5842&0.6467&0.7255\\\cdashline{2-12} 
            &LSPG(w/o ranking)&0.6385&0.3206&0.4619&0.5461&0.4382&0.3330&0.1996&0.8260&0.8858&0.9239\\ 
            &LSPG&\textbf{0.7119}&\textbf{0.3722}&\textbf{0.5123}&\textbf{0.5951}&\textbf{0.4983}&\textbf{0.3840}&\textbf{0.2275}&\textbf{0.8831}&\textbf{0.9184}&\textbf{0.9402}\\\hline

           \multirow{6}{*}{Portuguese}&GPT3(Ensemble)&\textbf{0.7700}&0.4358&0.5347&0.6229&0.5014&0.3620&0.2167&0.9171&\textbf{0.9491}&\textbf{0.9786}\\ 
          &GPT3(Single)&0.6363&0.3716&0.4615&0.5160&0.4105&0.2889&0.1615&0.7860&0.8181&0.8422\\\cdashline{2-12}
            &LSBERT&0.3262&0.1577&0.2326&0.2860&0.1904&0.1313&0.0775&0.4946&0.5802&0.6737\\
            &GMU-WLV&0.4812&0.2540&0.3716&0.3957&0.2816&0.1966&0.1153&0.6871&0.7566&0.8395\\\cdashline{2-12} 
            &LSPG(w/o ranking)&0.6176&0.3582&0.4839&0.5962&0.4135&0.3100&0.1899&0.8877&0.9278&0.9545\\
            &LSPG&0.7433&\textbf{0.4598}&\textbf{0.5989}&\textbf{0.6524}&\textbf{0.5023}&\textbf{0.3739}&\textbf{0.2250}&\textbf{0.9197}&\textbf{0.9491}&0.9625\\\hline        

    \end{tabular}
}
    \caption{Evaluation Results on English, Spanish and Portuguese languages. Only GPT3(Ensemble) over six different GPT3 prompts/configuration is few-shot method, and other methods are unsupervised or zero-shot methods. LSPG is our proposed method, and LSPG(w/o ranking) indicates that LSPG without the step of substitution ranking. Best values are bolded. }
    \label{tab:all_track} 
\end{table*}

\textbf{Metrics:} We use the same metrics with TSAR2022 shared task to evaluate the performance of LS methods for the three languages: ACC@1, MAP@$k$, Potential@$k$, Accuracy@$n$@top1 where $k$ $\in \{3, 5, 10\}$ and $n$ $\in \{1, 2, 3\}$. Potential@$k$ is defined at least one of the $k$ top-ranked substitutes is also present in the gold data. Accuracy@$k$@top1 evaluates whether most frequent suggested synonym in the gold data is also still in the generated candidates. MAP@$k$ additionally takes into account the position of the relevant substitutes among the first $k$ generated candidates. ACC@1 is same as Potential@1 and MAP@1. The above metrics account for various  aspects of methods' performances, allowing fair comparisons for different languages.

\textbf{Baselines:} We mainly compared our method LSPG with the following baselines.

(1) BERT-based LS methods. LSBERT\cite{qiang2021lsbert}, MANTIS\cite{li-etal-2022-mantis}, PresiUniv\cite{whistely2022presiuniv} and GMW-WLV\cite{north-etal-2022-gmu} are the most competitive BERT-based method in English, Spanish and Portuguese from TSAR-2022, respectively.

(2) GPT3-based LS methods \cite{aumiller-gertz-2022-unihd}. GPT3(Single) is a zero-shot prompted GPT-3 with a prompt asking for simplified synonyms given a particular context. GPT3(Ensemble) is an ensemble over six different GPT3 prompts/configurations with average rank aggregation. The version of GPT-3
used is text-davinci-002.

\textbf{Implementation details:} We employed Transformers\footnote{https://github.com/huggingface/transformers} for the implementation of our decoding method. The multilingual NMT we used is released with NLLB with 3.3B parameters which supports more than 200 languages\cite{costa2022no}. For English, the weights for prediction, frequency, word similarity are 0.04, 0.04, 0.1, respectively. For Spanish, the weights are 0.04,0.02, 0.4. For Portuguese, the weights are 0.04, 0.04, 0.4.  We finetune these hyper-parameters on the valid set separately. The number of output paraphrases $K$ is set to 50. The estimated suffix length during decoding is 3. We select up to 10 candidates for final evaluation.

\subsection{Experiment Results}

The results of our methods as well as the state-of-the-art methods are displayed in Table \ref{tab:all_track}. Because BERT-based LS methods are based on open source pre-trained language models, and GPT3-based methods are based on commercial APIs, we will discuss and analyze them separately.

Compared with BERT-based LS methods (LSBERT, MANTIS, PresiUniv, GMU-WLV), we can draw these conclusions. 

(1) Our method significantly outperforms BERT-based methods in all three languages. Even without the step of substitute ranking, our method LSPG(w/o ranking) is superior to the best BERT-based methods. 

(2) We see that BERT-based methods achieve better results in English than in Spanish and Portuguese languages. Unlike BERT-based methods which show significant performance gap between languages when utilizing separate pretrained models, our method exhibits stable performance across diverse languages using a solitary multilingual NMT.

Compared with GPT3-based methods, our method  improve the performance on all metrics, except ACC@1 and Potential@10 in Portuguese. Excluding performance advantages, our approach has the following advantages. 

(1) GPT3-based methods are only accessible through a paid interface and the best ensemble version even requires more than six visits to simplify a single complex word. They incur high costs in terms of both time and money to achieve satisfactory performance.  In contrast, our method is built on a freely available multilingual NMT, offering a more accessible and efficient solution. 

(2) GPT3(Ensemble) is a few-shot method, and LSPG is completely zero-shot. The performance of GPT3(Ensemble) is influenced by the provided demonstrations. For instance, on Spanish, our method significantly enhances the ACC@1 score from 0.6521 to 0.7119. 

(3) The experimental results of GPT3 are arduous to replicate, hence unfavorably impacting their potential as comparative results in the future. We will open source all of our codes and data results, in order to foster advancement in this domain.

Overall, our method has achieved a new state-of-the-art performance in multilingual LS, surpassing previous benchmarks in this field. It is clear that the success of our method can be attributed to our method of extracting substitutions during paraphrase decoding, which ensures that the original sentence's meaning is kept.

\subsection{Ablation Study}
To further analyze the factors affecting our method, we do more experiments in this subsection.

\begin{table}
    \centering
    \resizebox{\linewidth}{!}{
    \begin{tabular}{l|c|ccc|ccc}\hline
          &&\multicolumn{3}{c|}{Acc@k@Top1}&\multicolumn{3}{c}{MAP@k}\\\hline
          Feature&ACC@1&k=1&k=2&k=3&k=3&k=5&k=10\\\hline
            -&0.76&0.40&0.57&0.65&0.56&0.44&0.28\\          
            +freq&0.80&0.46&0.62&0.69&0.57&0.45&0.29\\
            +embed&0.75&0.40&0.55&0.66&0.57&0.44&0.29\\
            +both&\textbf{0.82}&\textbf{0.46}&\textbf{0.62}&\textbf{0.69}&\textbf{0.59}&\textbf{0.46}&\textbf{0.30}\\\hline
    \end{tabular}
    }
    \caption{Ablation study of ranking features for LSPG on English dataset. "+freq" indicates the results of adding only the frequency feature,  "+embed" indicates the results of adding only the embedding similarity feature, and "+both" is the proposed LSPG. Due to space limitation, we omit the results of little-varied Potential@k.}
    \label{tab:english track}
\end{table}
\begin{table*}[!h]
\centering

\begin{tabular}{l|l}
\hline
\textbf{Inst. 1} & Google says authorities in China have approved its \textbf{acquisition} of Motorola Mobility. \\\hline
Labels  &possession;purchase;buying;takeover;purchasing;gain;possesion;investment\\\hline
 GPT3(Ensemble) &\textcolor{red}{purchase};\textcolor{red}{takeover};\textcolor{red}{merger};obtaining;sale;procuring;\textcolor{blue}{buyout};consolidation;securing;picking up\\
  GPT3(Single) &\textcolor{red}{purchase};\textcolor{red}{takeover};merger;consolidation;\textcolor{blue}{buyout};procurement;obtainment;receipt;accession\\
 LSPG &\textcolor{red}{purchase};\textcolor{red}{takeover};acquiring;sale;buy;\textcolor{red}{buying};procurement;capture;\textcolor{red}{gain};merger\\
\hline

\textbf{Inst. 2} &  Nine people were reportedly killed in the \textbf{bombardment}. \\ \hline
Labels  &attack;strafe;shelling;burst;assault;bombing;incident;missile attack \\
\hline
 GPT3(Ensemble) & \textcolor{red}{shelling};barrage;\textcolor{red}{bombing};\textcolor{red}{attack};\textcolor{blue}{air strike};cannonade;fusillade;aerial bombardment;volleys;missile strike\\
 GPT3(Single)&\textcolor{blue}{air strike};missile strike;artillery strike;aerial bombardment;\textcolor{red}{bombing};\textcolor{red}{shelling};mortar fire;howitzer fire;cannon fire\\
LSPG&\textcolor{red}{bombing};\textcolor{red}{shelling};\textcolor{red}{attack};strike;fire;raid;shooting;blast;\textcolor{red}{assault};attacks\\
\hline
\textbf{Inst. 3} &The shootings are the worst \textbf{homicides} to take place in Brevard County since 1987.\\\hline
Labels&killings;murders;deaths\\
\hline
 GPT3(Ensemble) &\textcolor{red}{murders};\textcolor{red}{killings};executions;assassinations;\textcolor{blue}{slayings};shootings;stabbings;bloodbaths;massacres;rapes\\
 GPT3(Single)&\textcolor{red}{murders};\textcolor{red}{killings};shootings;stabbings;assaults;rapes;robberies;kidnappings;child abuse;domestic violence\\
 LSPG & \textcolor{red}{murders};\textcolor{red}{killings};murder;\textcolor{blue}{crimes};killing;cases;\textcolor{red}{deaths};shootings;crime;assassinations\\
\hline
\textbf{Inst. 4} &Six of the \textbf{ringleaders} have been captured and sent to other facilities.\\\hline
Labels&leaders;masterminds;bosses;instigators;troublemakers;captains;agitators\\
\hline
 GPT3(Ensemble) &\textcolor{red}{leaders};organizers;directors;\textcolor{red}{instigators};coordinators;\textcolor{red}{masterminds};\textcolor{blue}{chiefs};pioneers;executives;head honchos\\
 GPT3(Single)&\textcolor{red}{leaders};head honchos;bigwigs;top brass;heavyweights;power players;a-listers;big cheeses;big shots;big guns\\
 LSPG &\textcolor{red}{leaders};leader;leadership;\textcolor{red}{captains};\textcolor{blue}{heads};bandits;leads;conspirators;\textcolor{blue}{chiefs};members\\
\hline
\textbf{Inst. 5} &  The lowest \textbf{estimate} from Médecins Sans Frontières (MSF) is of 50 dead.\\\hline
Labels&estimation;assessment;figure;evaluation;reckon;guess;rough calculation\\
\hline
GPT3(Ensemble)&prediction;forecast;calculation;\textcolor{red}{guess};\textcolor{red}{estimation};\textcolor{blue}{guesstimate};projection;inference;opinion;surmise\\
GPT3(Single)&\textcolor{red}{guess};approximation;forecast;prediction;inference;conjecture;\textcolor{red}{estimation};supposition;surmise;postulation\\
 LSPG&\textcolor{red}{figure};\textcolor{red}{guess};\textcolor{red}{assessment};report;number;one;value;count;\textcolor{red}{evaluation};\textcolor{blue}{calculation}\\
\hline
\end{tabular}

\caption{The top 10 substitutes of five instances in English track of TSAR2022 shared task. The complex word is bolded, the substitutes in labels are marked in red and the suitable substitutions not in labels are in blue. }
\label{sample1}
\end{table*}

\textbf{Influence of ranking feature}
We add two new features (Frequency and Similarity) to rank the generated substitutions in our experiments. To further evaluate the effect of each feature on the final performance, we conduct an ablation study. Table 3 shows the results on the English dataset. It indicates that both frequency and embedding similarity features are beneficial to improve the performance of our method in various metrics, especially the frequency feature. This is becuase incorporating frequency score into the ranking step enables us to select simpler candidates.

\begin{table}
    \centering
    \resizebox{\linewidth}{!}{
    \begin{tabular}{l|c|ccc|ccc}\hline
          &&\multicolumn{3}{c|}{Acc@k@Top1}&\multicolumn{3}{c}{MAP@k}\\\hline
          Size&ACC@1&k=1&k=2&k=3&k=3&k=5&k=10\\\hline
            0.6B&0.73&0.39&0.55&0.62&0.54&0.41&0.26\\
            1.3B&0.76&\textbf{0.42}&0.55&0.62&0.56&0.44&0.27\\
            3.3B&\textbf{0.76}&0.40&\textbf{0.57}&\textbf{0.65}&\textbf{0.57}&\textbf{0.44}&\textbf{0.28}\\\hline
            % Para&0.71&0.36&0.54&0.64&0.52&0.40&0.25\\\hline
        
    \end{tabular}
    }
    \caption{Effect of varying different model size for LSPG on English dataset. We omit the results of little-varied Potential@k}
    \label{tab:abalation english track}
\end{table}

\textbf{Influence of model size:} The multilingual neural machine translation model we employ offers four variants with varying parameter scales (0.6B, 1.3B, 3.3B, and 55.3B), of which the default version in our experiments is 3.3B. To examine the impact of model size, we compare the performance of the first three models, as hardware limitations prevent us from assessing the 55.3B model. In Table 5, we report solely the model's predictive score to eliminate the influence of other ranking features. As anticipated, the 3.3B model yields the best performance, albeit with a marginal difference. To a certain extent, the increase in model parameters leads to improved simplification performance.

\begin{figure}[] %插入图片
\subfigure[]{
\pgfplotsset{compat=1.3}
\begin{tikzpicture}[scale=0.46,baseline] %tikz图片
\begin{axis}[
    xlabel=Estimated Suffix Length,
    ymin=0.5,
    ymax=1,
    ylabel=$Acc@1$,
    label style={font=\large},
    tick align=outside, %刻度在外显式
    legend style={at={(0.7,0.95)},anchor=north,font=\normalsize} 
    ]

\addplot[sharp plot,mark=*,blue,densely dotted] plot coordinates { 
    (0,0.65)
    (1,0.75)
    (2,0.76)
    (3,0.79)
    (4,0.76)
    (5,0.75)
};
\addlegendentry{English}
\addplot[sharp plot,mark=triangle,red,densely dotted] plot coordinates { 
    (0,0.60)
    (1,0.64)
    (2,0.64)
    (3,0.64)
    (4,0.64)
    (5,0.63)
};
\addlegendentry{Spanish}
\addplot[sharp plot,mark=square,orange] plot coordinates {
    (0,0.58)
    (1,0.60)
    (2,0.62)
    (3,0.62)
    (4,0.61)
    (5,0.61)
};
\addlegendentry{Portuguese}
\end{axis}
\end{tikzpicture}}
\subfigure[]{
\pgfplotsset{compat=1.3}
\begin{tikzpicture}[scale=0.46,baseline] %tikz图片
\centering
\begin{axis}[
    xlabel=Estimated Suffix Length, %横坐标名
    ymin=0,
    ymax=0.8,
    ylabel=$ACC@1@Top1$, %纵坐标名
    label style={font=\large},
    tick align=outside, %刻度在外显式
    legend style={at={(0.7,0.95)},anchor=north,{font=\normalsize}} %图例在图下方显示
    ]
\addplot[sharp plot,mark=*,blue,densely dotted] plot coordinates { 
    (0,0.34)
    (1,0.40)
    (2,0.40)
    (3,0.40)
    (4,0.40)
    (5,0.40)
};
\addlegendentry{English}
\addplot[sharp plot,mark=triangle,red] plot coordinates {
    (0,0.31)
    (1,0.32)
    (2,0.32)
    (3,0.32)
    (4,0.32)
    (5,0.32)
};
\addlegendentry{Spanish}
\addplot[sharp plot,mark=square,orange] plot coordinates {
    (0,0.34)
    (1,0.34)
    (2,0.35)
    (3,0.36)
    (4,0.35)
    (5,0.36)
};
\addlegendentry{Portuguese}
\end{axis}
\end{tikzpicture}}

\caption{Effect of varying estimated suffix length for LSPG. (a) the results on metric ACC@1, and (b) the results on metric Acc@1@Top1.} 
\label{fig:lookahed}
\end{figure}
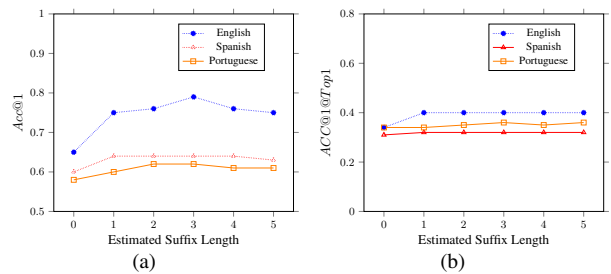

\textbf{Influence of length of estimated suffix:} Within our decoding process, we introduce a suffix estimate into the scoring function of our substitutions. This study examines the effect of the estimated suffix length, with Figure \ref{fig:lookahed} displaying the results. Ranging from 0 to 5, we manipulate the length and only record the predictive score to negate other factors' influence. Our findings indicate that estimating two or three suffix words produces optimal results, with no necessity to include further computations.

\subsection{Case Study}

In this case study, we compare our output substitutions against those generated by GPT3-based methods\cite{aumiller-gertz-2022-unihd}. Table 4 show five examples on the English track. Upon manual inspection, we found that both GPT3-based methods and our method generate suitable substitutions that are not included in labels. In general, Our method's actual performance is competitive with GPT(Ensemble) and outperforms GPT(Single). Notably, GPT3(Ensemble) requires combining the results of over six different GPT-3 prompts, including few-shot prompts, which is a time-consuming and costly process. In contrast, our method is completely zero-shot, utilizing a freely available multilingual NMT.

\section{Conclusion}
In this paper, we present a novel multilingual LS via paraphrase generation for generating meaning-preserved substitutions across multiple languages. We adopt a multilingual NMT system as the paraphraser that supports hundreds of languages. To address the challenges of identifying substitutions, we introduced a new decoding method that focuses on the lexical variations of the complex word. Our experiments demonstrate that our method achieves state-of-the-art results on the latest multilingual LS benchmarks, outperforming previous BERT-based approaches and showing competitive performance compared to ensemble GPT3-based method. We believe that our approach holds promise for a variety of natural language processing tasks, including but not limited to writing assistance and synonym extraction. Furthermore, our method is especially advantageous for low-resource languages.
\section*{Acknowledgement}
This research is partially supported by the National Natural Science Foundation of China under grants 62076217 and 61906060, and the Blue Project of Yangzhou University.
\bibliography{ref}

\begin{thebibliography}{10}

\bibitem{aumiller-gertz-2022-unihd}
Dennis Aumiller and Michael Gertz, `{U}ni{HD} at {TSAR}-2022 shared task: Is
  compute all we need for lexical simplification?', in {\em TSAR}, pp.
  251--258, (2022).

\bibitem{biran2011putting}
Or~Biran, Samuel Brody, and No{\'e}mie Elhadad, `Putting it simply: a
  context-aware approach to lexical simplification', in {\em ACL}, pp.
  496--501, (2011).

\bibitem{carroll1998practical}
John Carroll, Guido Minnen, Yvonne Canning, Siobhan Devlin, and John Tait,
  `Practical simplification of english newspaper text to assist aphasic
  readers', in {\em AAAI Workshop}, pp. 7--10, (1998).

\bibitem{costa2022no}
Marta~R Costa-juss{\`a}, James Cross, Onur {\c{C}}elebi, Maha Elbayad, Kenneth
  Heafield, Kevin Heffernan, Elahe Kalbassi, Janice Lam, Daniel Licht, Jean
  Maillard, et~al., `No language left behind: Scaling human-centered machine
  translation', {\em arXiv preprint arXiv:2207.04672}, (2022).

\bibitem{dabre2020survey}
Raj Dabre, Chenhui Chu, and Anoop Kunchukuttan, `A survey of multilingual
  neural machine translation', {\em ACM Computing Surveys (CSUR)}, {\bf 53}(5),
   1--38, (2020).

\bibitem{fan2021beyond}
Angela Fan, Shruti Bhosale, Holger Schwenk, Zhiyi Ma, Ahmed El-Kishky,
  Siddharth Goyal, Mandeep Baines, Onur Celebi, Guillaume Wenzek, Vishrav
  Chaudhary, et~al., `Beyond english-centric multilingual machine translation',
  {\em The Journal of Machine Learning Research}, {\bf 22}(1),  4839--4886,
  (2021).

\bibitem{finnimore2019strong}
Pierre Finnimore, Elisabeth Fritzsch, Daniel King, Alison Sneyd, Aneeq~Ur
  Rehman, Fernando Alva-Manchego, and Andreas Vlachos, `Strong baselines for
  complex word identification across multiple languages', in {\em NAACL}, pp.
  970--977, (2019).

\bibitem{Ganitkevitch2013}
Juri Ganitkevitch, Benjamin~Van Durme, and Chris Callison-Burch, `Ppdb: The
  paraphrase database', in {\em NAACL-HLT}, pp. 758--764, (2013).

\bibitem{glavavs2015simplifying}
Goran Glava{\v{s}} and Sanja {\v{S}}tajner, `Simplifying lexical
  simplification: Do we need simplified corpora?', in {\em ACL}, pp. 63--68,
  (2015).

\bibitem{gu-etal-2020-token}
Shuhao Gu, Jinchao Zhang, Fandong Meng, Yang Feng, Wanying Xie, Jie Zhou, and
  Dong Yu, `Token-level adaptive training for neural machine translation', in
  {\em EMNLP}, pp. 1035--1046, (2020).

\bibitem{hao2022parazh}
Wenjie Hao, Hongfei Xu, Deyi Xiong, Hongying Zan, and Lingling Mu, `Parazh-22m:
  A large-scale chinese parabank via machine translation', in {\em Proceedings
  of the 29th International Conference on Computational Linguistics}, pp.
  3885--3897, (2022).

\bibitem{hu2019parabank}
J~Edward Hu, Rachel Rudinger, Matt Post, and Benjamin Van~Durme, `Parabank:
  Monolingual bitext generation and sentential paraphrasing via
  lexically-constrained neural machine translation', in {\em AAAI}, volume~33,
  pp. 6521--6528, (2019).

\bibitem{hu-etal-2019-large}
J.~Edward Hu, Abhinav Singh, Nils Holzenberger, Matt Post, and Benjamin
  Van~Durme, `Large-scale, diverse, paraphrastic bitexts via sampling and
  clustering', in {\em CoNLL}, pp. 44--54, (2019).

\bibitem{jiang2019improving}
Shaojie Jiang, Pengjie Ren, Christof Monz, and Maarten de~Rijke, `Improving
  neural response diversity with frequency-aware cross-entropy loss', in {\em
  The World Wide Web Conference}, pp. 2879--2885, (2019).

\bibitem{johnson2017google}
Melvin Johnson, Mike Schuster, Quoc~V Le, Maxim Krikun, Yonghui Wu, Zhifeng
  Chen, Nikhil Thorat, Fernanda Vi{\'e}gas, Martin Wattenberg, Greg Corrado,
  et~al., `Google’s multilingual neural machine translation system: Enabling
  zero-shot translation', {\em Transactions of the Association for
  Computational Linguistics}, {\bf 5},  339--351, (2017).

\bibitem{Kriz2018Simplification}
Reno Kriz, Eleni Miltsakaki, Marianna Apidianaki, and Chris Callisonburch,
  `Simplification using paraphrases and context-based lexical substitution', in
  {\em NAACL}, pp. 207--217, (2018).

\bibitem{li-etal-2022-mantis}
Xiaofei Li, Daniel Wiechmann, Yu~Qiao, and Elma Kerz, `{MANTIS} at {TSAR}-2022
  shared task: Improved unsupervised lexical simplification with pretrained
  encoders', in {\em TSAR}, pp. 243--250, (2022).

\bibitem{lin-etal-2022-improving}
Yu~Lin, Zhecheng An, Peihao Wu, and Zejun Ma, `Improving contextual
  representation with gloss regularized pre-training', in {\em NAACL}, pp.
  907--920, (2022).

\bibitem{lu-etal-2022-neurologic}
Ximing Lu, Sean Welleck, Peter West, Liwei Jiang, Jungo Kasai, Daniel Khashabi,
  Ronan Le~Bras, Lianhui Qin, Youngjae Yu, Rowan Zellers, Noah~A. Smith, and
  Yejin Choi, `{N}euro{L}ogic a*esque decoding: Constrained text generation
  with lookahead heuristics', in {\em Proceedings of the 2022 Conference of the
  North American Chapter of the Association for Computational Linguistics:
  Human Language Technologies}, pp. 780--799, Seattle, United States, (July
  2022). Association for Computational Linguistics.

\bibitem{north-etal-2022-gmu}
Kai North, Alphaeus Dmonte, Tharindu Ranasinghe, and Marcos Zampieri,
  `{GMU}-{WLV} at {TSAR}-2022 shared task: Evaluating lexical simplification
  models', in {\em Proceedings of the Workshop on Text Simplification,
  Accessibility, and Readability (TSAR-2022)}, pp. 264--270, Abu Dhabi, United
  Arab Emirates (Virtual), (December 2022). Association for Computational
  Linguistics.

\bibitem{paetzold2016unsupervised}
Gustavo Paetzold and Lucia Specia, `Unsupervised lexical simplification for
  non-native speakers', in {\em AAAI}, volume~30, (2016).

\bibitem{paetzold2017survey}
Gustavo~H Paetzold and Lucia Specia, `A survey on lexical simplification', {\em
  Journal of Artificial Intelligence Research}, {\bf 60},  549--593, (2017).

\bibitem{pavlick2016simple}
Ellie Pavlick and Chris Callison-Burch, `Simple ppdb: A paraphrase database for
  simplification', in {\em ACL}, pp. 143--148, (2016).

\bibitem{pavlick2015ppdb}
Ellie Pavlick, Pushpendre Rastogi, Juri Ganitkevitch, Benjamin Van~Durme, and
  Chris Callison-Burch, `Ppdb 2.0: Better paraphrase ranking, fine-grained
  entailment relations, word embeddings, and style classification', in {\em
  ACL}, pp. 425--430, (2015).

\bibitem{qiang2021lsbert}
Jipeng Qiang, Yun Li, Yi~Zhu, Yunhao Yuan, Yang Shi, and Xindong Wu, `Lsbert:
  Lexical simplification based on bert', {\em IEEE/ACM Transactions on Audio,
  Speech, and Language Processing}, {\bf 29},  3064--3076, (2021).

\bibitem{qiang2020AAAI}
Jipeng Qiang, Yun Li, Yi~Zhu, Yunhao Yuan, and Xindong Wu, `Lexical
  simplification with pretrained encoders', {\em Thirty-Fourth AAAI Conference
  on Artificial Intelligence},  8649–8656, (2020).

\bibitem{qiang2021chinese}
Jipeng Qiang, Xinyu Lu, Yun Li, Yunhao Yuan, and Xindong Wu, `Chinese lexical
  simplification', {\em IEEE/ACM Transactions on Audio, Speech, and Language
  Processing}, {\bf 29},  1819--1828, (2021).

\bibitem{qiang2021unsupervised}
Jipeng Qiang and Xindong Wu, `Unsupervised statistical text simplification',
  {\em IEEE Transactions on Knowledge and Data Engineering}, {\bf 33}(4),
  1802--1806, (2021).

\bibitem{qiang2023natural}
Jipeng Qiang, Shiyu Zhu, Yun Li, Yi~Zhu, Yunhao Yuan, and Xindong Wu, `Natural
  language watermarking via paraphraser-based lexical substitution', {\em
  Artificial Intelligence},  103859, (2023).

\bibitem{saggion2017automatic}
Horacio Saggion, `Automatic text simplification', {\em Synthesis Lectures on
  Human Language Technologies}, {\bf 10}(1),  1--137, (2017).

\bibitem{saggion2023findings}
Horacio Saggion, Sanja {\v{S}}tajner, Daniel Ferr{\'e}s, Kim~Cheng Sheang,
  Matthew Shardlow, Kai North, and Marcos Zampieri, `Findings of the tsar-2022
  shared task on multilingual lexical simplification', {\em arXiv preprint
  arXiv:2302.02888}, (2023).

\bibitem{vstajner2022lexical}
Sanja {\v{S}}tajner, Daniel Ferr{\'e}s, Matthew Shardlow, Kai North, Marcos
  Zampieri, and Horacio Saggion, `Lexical simplification benchmarks for
  english, portuguese, and spanish', {\em Frontiers in artificial
  intelligence}, {\bf 5},  991242, (2022).

\bibitem{tan-etal-2019-multilingual}
Xu~Tan, Jiale Chen, Di~He, Yingce Xia, Tao Qin, and Tie-Yan Liu, `Multilingual
  neural machine translation with language clustering', in {\em Proceedings of
  the 2019 Conference on Empirical Methods in Natural Language Processing and
  the 9th International Joint Conference on Natural Language Processing
  (EMNLP-IJCNLP)}, pp. 963--973, Hong Kong, China, (November 2019). Association
  for Computational Linguistics.

\bibitem{thompson2020paraphrase}
Brian Thompson and Matt Post, `Paraphrase generation as zero-shot multilingual
  translation: Disentangling semantic similarity from lexical and syntactic
  diversity', in {\em Proceedings of the Fifth Conference on Machine
  Translation}, pp. 561--570, (2020).

\bibitem{vamvas-sennrich-2022-nmtscore}
Jannis Vamvas and Rico Sennrich, `{NMTS}core: A multilingual analysis of
  translation-based text similarity measures', in {\em Findings of the
  Association for Computational Linguistics: EMNLP 2022}, pp. 198--213, Abu
  Dhabi, United Arab Emirates, (December 2022). Association for Computational
  Linguistics.

\bibitem{vijayakumar2018diverse}
Ashwin Vijayakumar, Michael Cogswell, Ramprasaath Selvaraju, Qing Sun, Stefan
  Lee, David Crandall, and Dhruv Batra, `Diverse beam search for improved
  description of complex scenes', in {\em Proceedings of the AAAI Conference on
  Artificial Intelligence}, volume~32, (2018).

\bibitem{wada-etal-2022-unsupervised}
Takashi Wada, Timothy Baldwin, Yuji Matsumoto, and Jey~Han Lau, `Unsupervised
  lexical substitution with decontextualised embeddings', in {\em Proceedings
  of the 29th International Conference on Computational Linguistics}, pp.
  4172--4185, (2022).

\bibitem{whistely-etal-2022-presiuniv}
Peniel Whistely, Sandeep Mathias, and Galiveeti Poornima, `{P}resi{U}niv at
  {TSAR}-2022 shared task: Generation and ranking of simplification substitutes
  of complex words in multiple languages', in {\em Proceedings of the Workshop
  on Text Simplification, Accessibility, and Readability (TSAR-2022)}, pp.
  213--217, Abu Dhabi, United Arab Emirates (Virtual), (December 2022).
  Association for Computational Linguistics.

\bibitem{whistely2022presiuniv}
Peniel Whistely, Sandeep Mathias, and Galiveeti Poornima, `Presiuniv at
  tsar-2022 shared task: Generation and ranking of simplification substitutes
  of complex words in multiple languages', in {\em TSAR}, pp. 213--217, (2022).

\bibitem{wieting2017paranmt}
John Wieting and Kevin Gimpel, `Paranmt-50m: Pushing the limits of paraphrastic
  sentence embeddings with millions of machine translations', {\em arXiv
  preprint arXiv:1711.05732}, (2017).

\bibitem{zhou2021paraphrase}
Jianing Zhou and Suma Bhat, `Paraphrase generation: A survey of the state of
  the art', in {\em Proceedings of the 2021 Conference on Empirical Methods in
  Natural Language Processing}, pp. 5075--5086, (2021).

\end{thebibliography}
\end{document}